\documentclass[nohyperref]{article}
\usepackage{aas_macros}
\usepackage[accepted]{icml2022}
\usepackage{microtype}

\usepackage{graphicx}
\usepackage{subfigure}
\usepackage{booktabs} 

\usepackage{hyperref}
\usepackage{comment}
\usepackage{xcolor}
\usepackage{siunitx}
\usepackage{mathtools}

\usepackage{xspace}

\usepackage{dsfont}

\usepackage{amsfonts} 

\newcommand{\loss}{\mathcal{L}}
\newcommand{\x}{\mathbf{x}}
\newcommand{\y}{\mathbf{y}}

\usepackage{textcomp}

\usepackage{todonotes}

\usepackage{cleveref}

\graphicspath{{figures/}}

\newcommand{\tardiscode}{TARDIS\xspace}
\graphicspath{{figures/}}

\icmltitlerunning{Probabilistic Dalek}

\begin{document}
\title{Dalek II - Emulator framework with probabilistic inference}
\icmltitle{Probabilistic Dalek - Emulator framework with probabilistic prediction for supernova tomography}
\icmlsetsymbol{equal}{*}

\begin{icmlauthorlist}
    \icmlauthor{Wolfgang Kerzendorf}{equal,msuastro,msucmse}
    \icmlauthor{Nutan Chen}{equal,volkswagen}
    \icmlauthor{Jack O'Brien}{msuastro}
    \icmlauthor{Johannes Buchner}{mpe}
    \icmlauthor{Patrick van der Smagt}{volkswagen,budapest}
\end{icmlauthorlist}

\icmlaffiliation{msuastro}{Department of Physics and Astronomy, Michigan State University, East Lansing, MI 48824, USA}
\icmlaffiliation{msucmse}{Department of Computational Mathematics, Science, and Engineering, Michigan State University, East Lansing, MI 48824, USA}
\icmlaffiliation{mpe}{Max-Planck-Institut f\"{u}r extraterrestrische Physik, Giessenbachstrasse 1, 85748 Garching bei M\"{u}nchen, Germany}
\icmlaffiliation{volkswagen}{Machine Learning Research Lab, Volkswagen AG, Munich, Germany}
\icmlaffiliation{budapest}{Faculty of Informatics, E\"otv\"os Lor\'and University, Budapest, Hungary}

\icmlcorrespondingauthor{Wolfgang Kerzendorf}{wkerzend@msu.edu,wkerzendorf@gmail.com}

\printAffiliationsAndNotice{\icmlEqualContribution}


\begin{abstract}
Supernova spectral time series can be used to reconstruct a spatially resolved explosion model known as supernova tomography. In addition to an observed spectral time series, a supernova tomography requires a radiative transfer model to perform the inverse problem with uncertainty quantification for a reconstruction. The smallest parametrizations of supernova tomography models are roughly a dozen parameters with a realistic one requiring more than 100. Realistic radiative transfer models require tens of CPU minutes for a single evaluation making the problem computationally intractable with traditional means requiring millions of MCMC samples for such a problem. A new method for accelerating simulations known as surrogate models or emulators using machine learning techniques offers a solution for such problems and a way to understand progenitors/explosions from spectral time series. There exist emulators for the \tardiscode supernova radiative transfer code but they only perform well on simplistic low-dimensional models (roughly a dozen parameters) with a small number of applications for knowledge gain in the supernova field. In this work, we present a new emulator for the radiative transfer code \tardiscode that not only outperforms existing emulators but also provides uncertainties in its prediction. It offers the foundation for a future active-learning-based machinery that will be able to emulate very high dimensional spaces of hundreds of parameters crucial for unraveling urgent questions in supernovae and related fields. 
\end{abstract}


\section{Introduction}


Regular stars only allow for direct measurements of the properties of their surface and views into the interior are not directly possible. The dynamical nature of supernovae, however, encodes spatially resolved information about their interior in spectral time series (known as supernova tomography). Supernova tomography uses the recession of a surface of last scattering (photosphere) deeper into the envelope to perform parameter inference on shells of ejecta with theoretical radiative transfer simulations. Such data-driven physically motivated tomography models are directly comparable to explosion simulations and have the power to unlock the progenitor and explosion mechanisms for many different classes of exploding objects. 

However, the parameter space for such models has at a minimum $\approx 12$ dimensions for very simple parameterizations of the problem and coupled with evaluations times of tens of minutes per model with modern radiative transfer code results in a computationally intractable model \citep[see e.g.][]{kerzendorf2021dalek}. Credible comparisons of supernova tomographies with explosion scenarios do not only require solving the inverse problem but also demand uncertainty quantification exacerbating the computational requirements. 

Simulation based inference has addressed the uncertainty quantification by using emulators (also known as surrogate models) which are machine learning constructs provide approximations that are orders of magnitude faster  to complex simulations which then allows their use in standard MCMC samplers
\citep{cranmer_sbi2020}. \citet{refId0,kerzendorf2021dalek, obrien21} have developed emulators to accelerate the \tardiscode \citep{2014MNRAS.440..387K, 2019A&A...621A..29V, kerzendorf_wolfgang_2022_6527890} radiative transfer code for problems up to 14 parameters and applied them to supernova tomography. 

A full tomography and with it the understanding of explosion and progenitor mechanisms requires roughly an order of magnitude more parameters. Such large dimensionality is not feasible for the current generation of emulators. High-dimensional emulators will require an active learning component to ensure training samples are only produced in parts of the parameter space where they decrease the uncertainty of the emulator. 

Numerous publications extend neural networks to quantify the prediction by uncertainties. Recently popular approaches of Bayesian methods include Monte Carlo (MC) dropout \cite{gal2016dropout} and weight uncertainty \cite{blundell2015weight}. In addition, there are non-Bayesian approaches, for instance, post-hoc calibration by temperature scaling of a validation dataset \cite{guo2017calibration}, and deep ensembles \cite{NIPS2017_9ef2ed4b}. Amongst the classical methods, deep ensembles \cite{NIPS2017_9ef2ed4b} generally perform best in uncertainty estimation \cite{NEURIPS2019_8558cb40}. Methods such as weight uncertainty, MC dropout, and deep ensembles require multiple passes to obtain the uncertainty. For computational efficiency, deterministic methods in a single forward pass \cite{van2020uncertainty, NEURIPS2020_543e8374} are presented. 

Since our model is a relatively small network which does not take a large amount of computation, we choose the best uncertainty prediction, viz.\ deep ensembles.
In our case, the prediction uncertainties are not only driven by the sampling sparseness of the parameter space but also by the Monte Carlo nature of the \tardiscode radiative transfer code.

In \Cref{sec:neural_network_model}, we describe the setup of our neural network model. We show various statistics about our model and comparison with \tardiscode and its uncertainty in \Cref{sec:results}. We conclude and give an outlook over future work in \Cref{sec:conclusion}.

\section{Neural network model\label{sec:neural_network_model}}

The mapping from the parameters to the spectra is approbated by a feedforward neural network.
For uncertainty estimation, we select proper scoring rules, and then train the ensemble as proposed in \cite{NIPS2017_9ef2ed4b}.
An adversarial training (AT) is an option that probably improves the uncertainty measurement.

\subsection{Single model for regression}
\label{sec:single_model}

We use the training set from \citep{kerzendorf2021dalek} and thus have the `parameters' as the inputs $\x \in \mathbb{R}^{12}$ and the `spectra' as the outputs $\y \in \mathbb{R}^{500}$ of the neural networks. The dataset is split into \num{90000} training data, \num{18000} validation data and \num{18000} test data. 
The data is preprocessed by a logarithmic scale and then normalised by removing the mean and scaling to unit variance \citep{pedregosa2011scikit}.

The data sets are comfortably large.  That may be the reason that the neural networks are not very sensitive to hyperparameter variations, as we discovered in a very broad search on a compute cluster.  The use of dropout did not change much, nor was layer normalisation essential. The best architectures were those with between 3 and 5 hidden layers of 200 to 400 softplus hidden units, training with batch sizes between 100 and 500 samples.  All networks are trained with Adam.

\citet{van1998solving,he2016deep} propose a residual architecture for deep layers to avoid degradation problems. This residual architecture aggregates the output from the previous layer and the current layer as the input to the next layer. In our network, we use concatenation for faster convergence as opposed to addition \citep[as suggested in][]{he2016deep}.

We obtain the uncertainty for each model by outputting the mean $\mu$, and the standard deviation (STD) $\sigma$, from the final hidden layer.

\paragraph{Scoring rules}
The quality of predictive uncertainty can be measured by scoring rules. The maximised likelihood $\log p_{\theta} (\y \vert \x)$ is a proper scoring rule \cite{NIPS2017_9ef2ed4b,gneiting2007strictly}.
We use a maximum-a-posteriori (MAP) that is a negative log likelihood (NLL) with the weight regularisation $\log p(\theta)$. A standard NLL uses no prior knowledge about the expected distribution of the model weights $\theta$ and in our case leads to overfit. If the weights are standard distribution, $\log p(\theta)$ is approximate to $L_2$ norm of the weights. In our implementation, we set a hyper-parameter $\beta$ as the coefficient of the regularisation, and the loss is minimised as:

\begin{align}
\label{eq:loss}
\loss(\x) =& -\log p_\theta(\y\vert\x) - \beta \log p(\theta) \\
\propto &\frac{\log \sigma^2_\theta(\x)}{2} + \frac{\bigl(\y-\mu_\theta (\x)\bigr)^2}{2\sigma^2_\theta (\x)} + \beta \| \theta \|_2 + \mathrm{constant}. \nonumber 
\end{align}

\paragraph{Adversarial training}
An adversarial training \citep[AT][]{Szegedy, Goodfellow_2015} is able to smooth predictive distributions and is possible to improve the uncertainty prediction \cite{NIPS2017_9ef2ed4b}. 
Adversarial examples are similar to the training samples, but are misclassified by NNs. The examples are generated by $\x' = \x + \epsilon~\mathrm{sign}(\nabla _\x \loss (\theta, \x, \y))$, where the perturbation is bounded in $\epsilon$.

We then rewrite the loss:
\begin{align}
\loss_{AT}(\x) = \alpha\loss(\x) + (1-\alpha) \loss(\x').
\end{align}

\begin{table*}[]
\vskip -0.1in
\caption{Model architecture.}
\begin{center}
\begin{footnotesize}
\begin{sc}				
\begin{tabular}{lc}
\toprule
name & hyper-parameters \\
\midrule
input dimension & 12 \\
output dimension & 500 \\
number of hidden layers & 5\\
activation of hidden layers & Softplus\\
connection of hidden layers & residual with concatenation \\
hidden units & 400 \\
Activation of $\mu$ final layer & Linear\\
Activation of $\sigma$ final layer & Softplus\\
$\beta$ & 5e-4 \\
$\alpha$ &  0.9 \\
$\epsilon$ & 5e-4\\
batch size 	& 500 \\
learning rate & 2e-4\\
parameters initialisation & xavier normal \\
\bottomrule
\end{tabular}
\end{sc}
\end{footnotesize}
\end{center}
\vskip -0.1in
\label{table:architecture}
\end{table*}

We select a good model by searching for the hyper-parameters based on the lowest loss value of the validation dataset. 
Searching in small ranges around the best hyper-parameters of \cite{kerzendorf2021dalek}, we obtain \Cref{table:architecture} using Polyaxon 0.5.6 (\url{https://polyaxon.com}) on a cluster with multiple NVIDIA Tesla V100 GPUs. The code is implemented in Pytorch 1.7.1 \citep{NEURIPS2019_9015}.

\subsection{Ensembles}

To obtain multiple models from the best architecture, we have different weight initialisation and the order of the batch data selection, for each model.  
After training $M$ models independently and in parallel, the prediction is measured using a uniformly-weighted mixture of Gaussian distributions
\begin{align}
p(\y\vert \x) = M^{-1} \sum^M_{m=1} p_{\theta_m}(\y\vert\x, \theta_m).
\end{align}
We approximate the ensemble prediction as a Gaussian with a mean and variance equal to, respectively, the mean and variance of the mixture
\begin{align}
\mu_{\ast}(\x) = & M^{-1}\sum_m\mu_{\theta_m}(\x), \nonumber \\
\sigma^2_\ast(\x) =& M^{-1}\sum_m\bigl(\sigma^2_{\theta_m}(\x) + \mu^2_{\theta_m}(\x)\bigr) - \mu^2_\ast(\x).
\end{align}


\section{Results\label{sec:results}}

\begin{figure*}[ht!]
	\centering
	\includegraphics[width=\columnwidth]{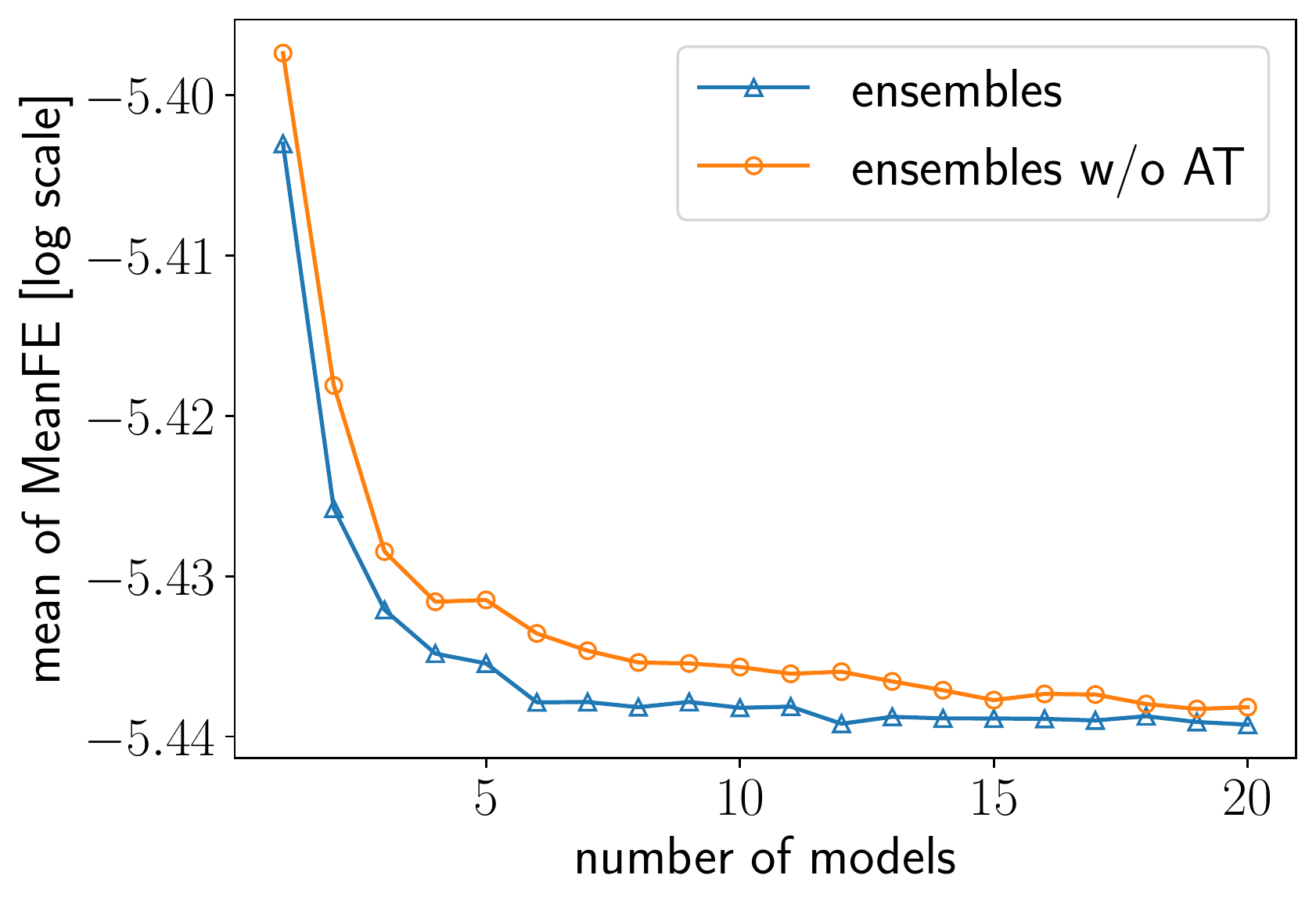}
	\includegraphics[width=\columnwidth]{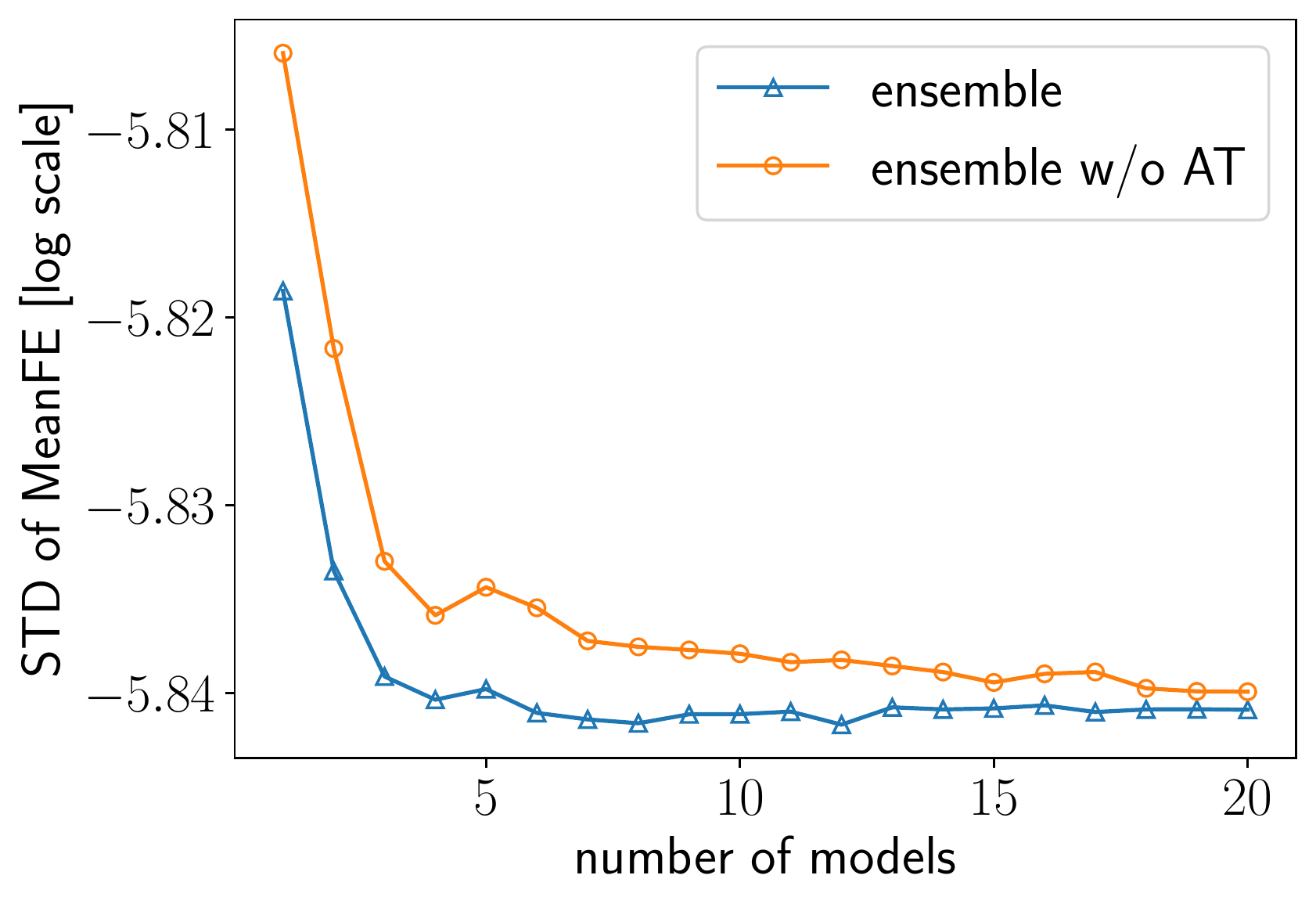}
	\caption{Accuracy. The horizontal-axis shows the number of NN in the ensembles. The mean and the STD of \num{18000} testing samples are computed. }%
	\label{fig:accuracy}
\end{figure*}

We evaluate the approach on the spectra simulation. 
We take the empirical variance as the baseline, which is commonly used in practice. Ensembles of NNs approximate the uncertainty from the empirical variance of multiple predictions. Usually, it uses mean square error (MSE) loss for the training. To simplify the comparison, we use the same models as the deep ensembles. We also compare the deep ensembles with a vanilla uncertainty estimation -- measuring the uncertainty by the STD of a single model. Since the vanilla approach is the deep ensembles with $M = 1$, we do not separately demonstrate the results.
Additionally, we evaluate how the optional term, the AT, affects the ensembles. 

We use the mean and max of fractional error metrics as in \cite{refId0, kerzendorf2021dalek} to quantify the results:
\begin{align}
\mathrm{MeanFE} =& \frac{1}{N}\sum^N_{i=0} \frac{| \y^\text{emu}_{i}-\y^\text{test}_{i}|}{\y^\text{test}_{i}} \nonumber, \\
\mathrm{MaxFE} =& \max^N_{i=0} \frac{| \y^\text{emu}_{i}-\y^\text{test}_{i}|}{\y^\text{test}_{i}}, \nonumber
\end{align}
where $N$ is the dimension of spectra, and $\y_i$ represents the flux at the $i$-th dimension.



\subsection{Spectra prediction}

\begin{figure*}[]
	\centering
	\includegraphics[width=0.45\textwidth]{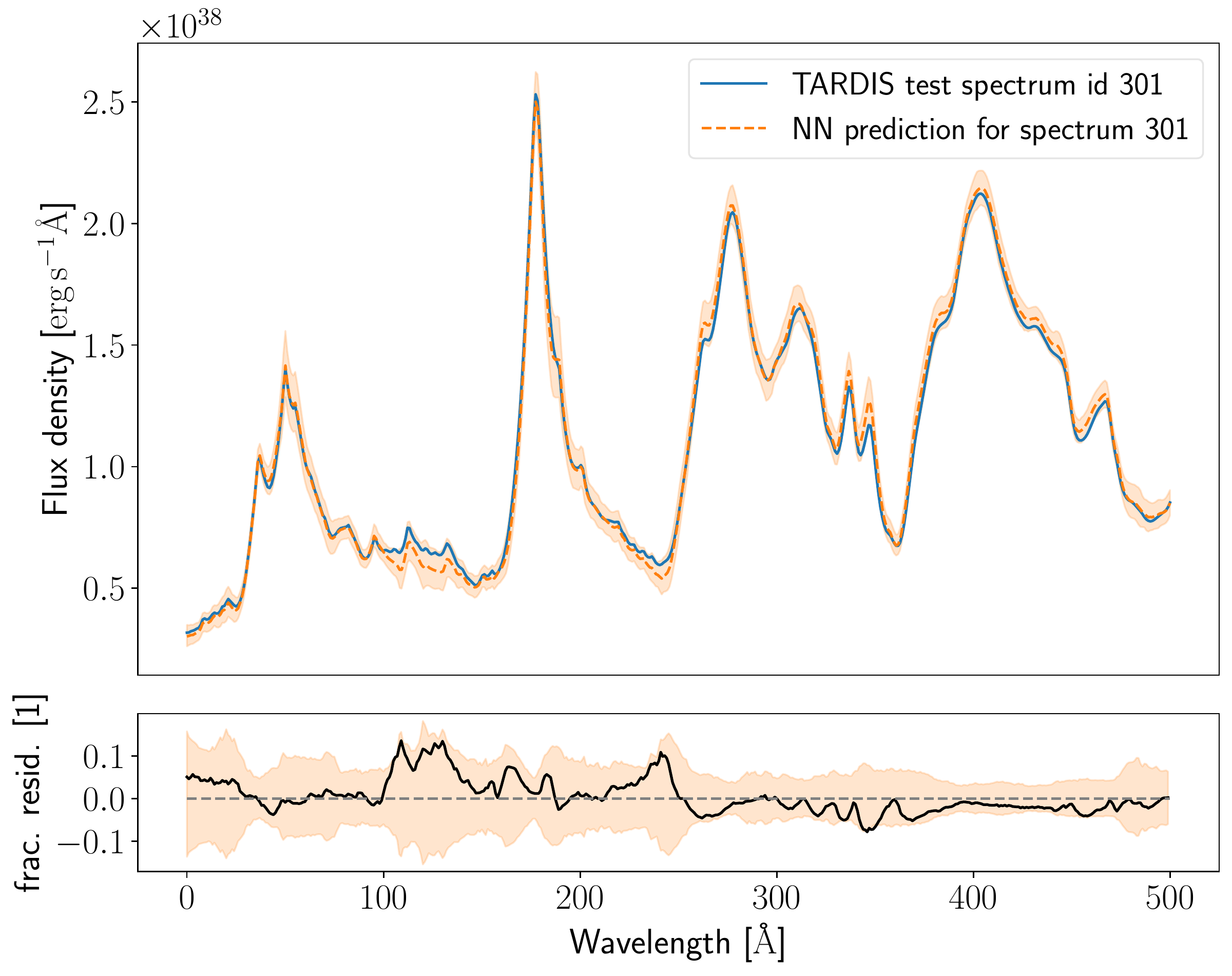}
	\includegraphics[width=0.45\textwidth]{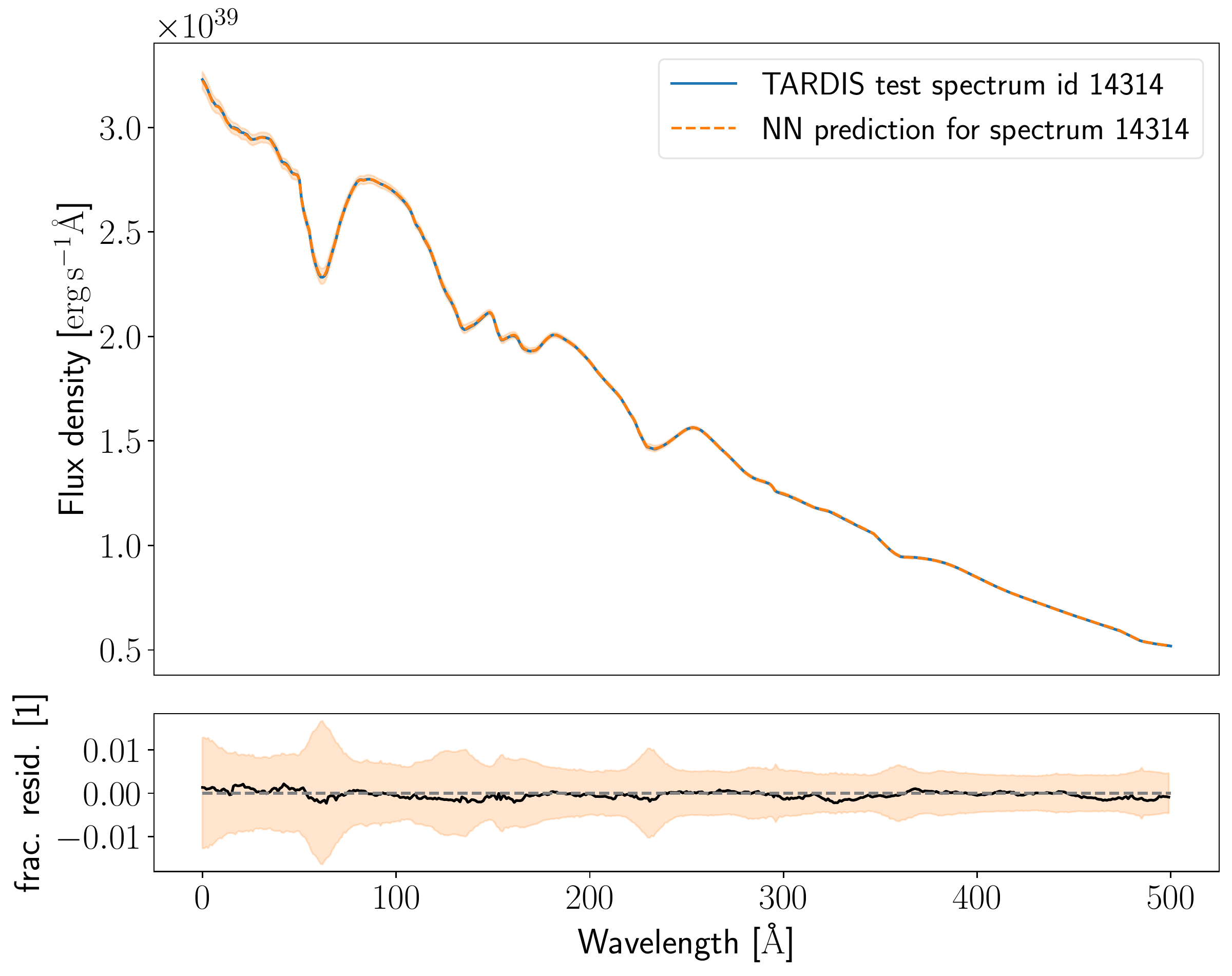}
	\caption{Examples of the uncertainty prediction.
	Highest and the lowest MaxFE from the test set predictions. 
	The shaded areas denote 99.9\,\% confidence interval. In the lower figures, the black lines and dashed grey lines represent the residual and zero values, respectively.}%
	\label{fig:prediction_frac_resid}
\end{figure*}

\Cref{fig:accuracy} shows the accuracy of the prediction. In general, the ensembles with ATs outperforms the approaches without ATs. With the number of models in the ensemble more than six, the accuracy is not significantly improved. 

Based on the above evaluation, the AT hardly improves the uncertainty prediction, but it improves the accuracy, especially when the number of models is small. The deep ensembles outperform the empirical and vanilla approaches since the latter is easily over-confident. Taking into the accuracy, computation efficiency, and the uncertainty prediction into consideration, the best number of the models for the deep ensembles is suggested to be six.
\Cref{fig:prediction_frac_resid} illustrates two examples of the prediction of the testing dataset using deep ensembles with six models and thus in the following we will use six models for the ensemble in our further tests.

\subsection{Uncertainty prediction}

\begin{figure}[]
	\centering
	\includegraphics[width=\columnwidth]{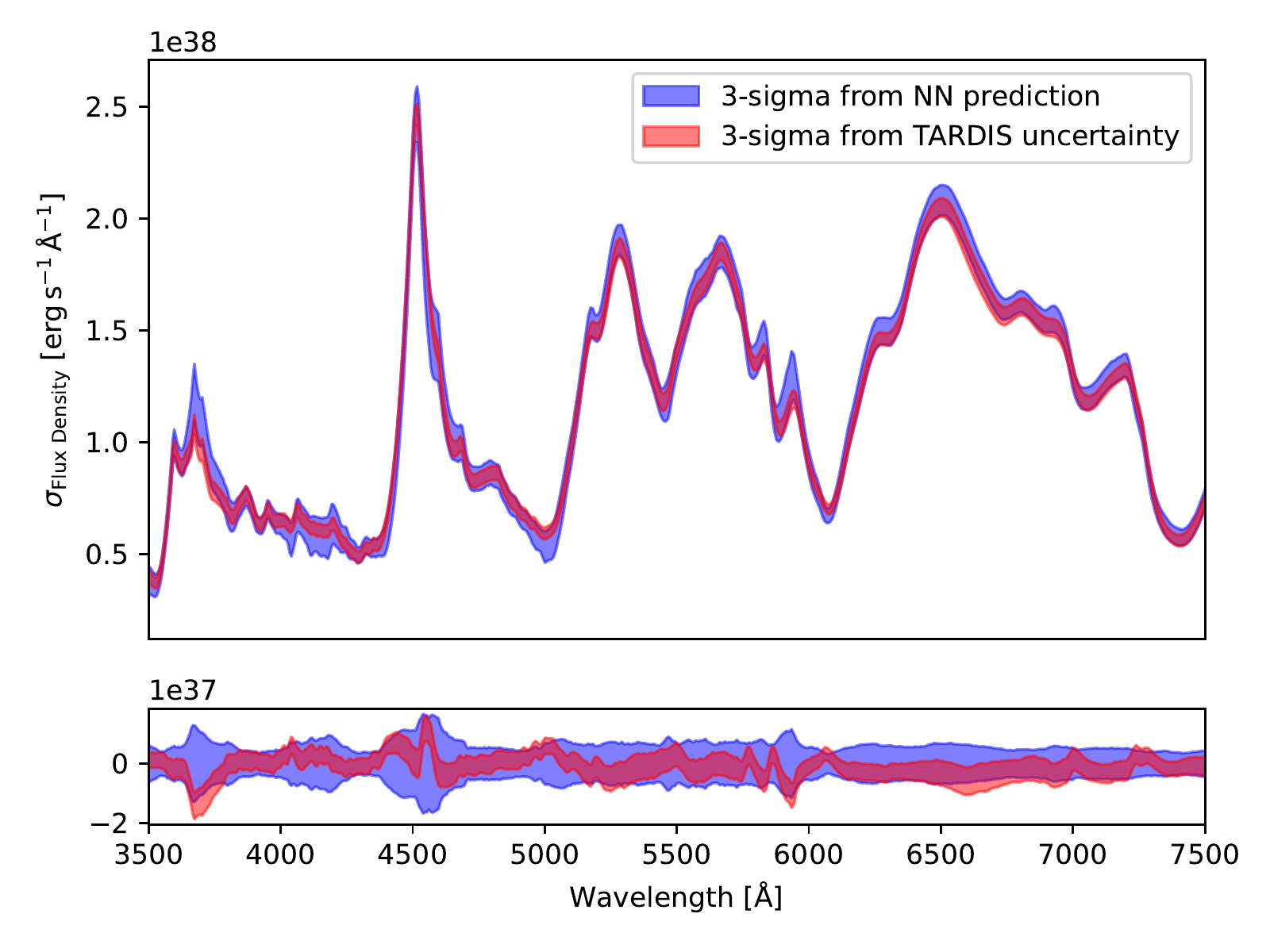}
	\caption{Comparison between the Monte Carlo uncertainty arising from \tardiscode and the uncertainty estimated by the deep ensemble for test set sample 12625. We show $3-\sigma$ for visualization purposes.}%
	\label{fig:uncertainty_mc}
\end{figure}

The \textsc{Dalek} emulator has to contend with two types of uncertainties: 1) the prediction uncertainty stemming from a sparseness of sampling 2) the intrinsic uncertainty of the \tardiscode simulator arising from the Monte Carlo radiative transfer. We have estimated the uncertainty given by \tardiscode for the test-set spectrum with the highest $\max\sigma/\mu$-ratio (highest relative uncertainty; id=12625) and reran \tardiscode (version hash \textrm{ad91bef1a}) with 100 different seeds to estimate the uncertainty arising from the Monte Carlo method. In \Cref{fig:uncertainty_mc}, we show that the network predicts a larger uncertainty that likely includes additional prediction uncertainty and mostly completely envelops the uncertainty given by \tardiscode. Further tests are needed with a larger number of samples to explore the consistency of this result.

\section{Conclusions\label{sec:conclusion}}

We present a probabilistic neural network model for the \tardiscode supernova radiative transfer code. The emulator model is based on the deep ensemble approach given by \citep{NIPS2017_9ef2ed4b} and even for a single model provides a MeanFE of $\approx 10^{-5}$ which is better than the model in the original \textsc{Dalek emulator} \citep[MeanFe $\approx 10^{-3}$; ][]{kerzendorf2021dalek}. In \Cref{fig:accuracy}, we show that for this architecture there is no substantial increase in accuracy with more than six models. We show that the test set \tardiscode spectrum lies well within the predicted uncertainties (see \Cref{fig:uncertainty_mc}). We also show that (for now for a limited set) the uncertainty predicted by the neural network also captures the Monte Carlo uncertainty by \tardiscode well. The preliminary work that is shown is very promising but still demands several more tests. As discussed in the introduction, the described work is part of a larger effort to build an active learning emulator for \tardiscode with the express goal of doing supernova tomography and will be explored in future work.
\begin{appendix}
\section{Contributor Role Taxonomy}
We will use the Contributor Role Taxonomy as outlined in \url{https://credit.niso.org/}.
\begin{itemize}
    \item Conceptualization: van der Smagt, Kerzendorf, Chen
    \item Formal Analysis: Chen, Kerzendorf
    \item Funding acquisition: van der Smagt, Kerzendorf
    \item Investigation: Chen, van der Smagt, Kerzendorf
    \item Methodology: Chen, van der Smagt
    \item Project administration: van der Smagt, Chen, Kerzendorf
    \item Software: Chen, Kerzendorf, van der Smagt, O'Brien
    \item Supervision: Kerzendorf, van der Smagt
    \item Writing - original draft: Chen, Kerzendorf
    \item Writing - review \& editing: Chen, Kerzendorf, Buchner
\end{itemize}
\section{Acknowledgements}
This research made use of TARDIS, a community-developed software package for spectral
synthesis in supernovae \citep{2014MNRAS.440..387K,kerzendorf_wolfgang_2022_6430631}. The
development of TARDIS received support from GitHub, the Google Summer of Code
initiative, and from ESA's Summer of Code in Space program. TARDIS is a fiscally
sponsored project of NumFOCUS. TARDIS makes extensive use of Astropy \citep{2013A&A...558A..33A,2018AJ....156..123A,2022arXiv220614220T}, pandas \citep{jeff_reback_2022_6702671}, and Numpy \citep{2020Natur.585..357H}.
\end{appendix}


\newpage

\bibliography{dalek}

\begin{thebibliography}{27}
\providecommand{\natexlab}[1]{#1}
\providecommand{\url}[1]{\texttt{#1}}
\expandafter\ifx\csname urlstyle\endcsname\relax
  \providecommand{\doi}[1]{doi: #1}\else
  \providecommand{\doi}{doi: \begingroup \urlstyle{rm}\Url}\fi

\bibitem[{Astropy Collaboration} et~al.(2013){Astropy Collaboration},
  Robitaille, Tollerud, Greenfield, Droettboom, Bray, Aldcroft, Davis,
  Ginsburg, {Price-Whelan}, Kerzendorf, Conley, Crighton, Barbary, Muna,
  Ferguson, Grollier, Parikh, Nair, Unther, Deil, Woillez, Conseil, Kramer,
  Turner, Singer, Fox, Weaver, Zabalza, Edwards, Azalee~Bostroem, Burke, Casey,
  Crawford, Dencheva, Ely, Jenness, Labrie, Lim, Pierfederici, Pontzen, Ptak,
  Refsdal, Servillat, and Streicher]{2013A&A...558A..33A}
{Astropy Collaboration}, Robitaille, T.~P., Tollerud, E.~J., Greenfield, P.,
  Droettboom, M., Bray, E., Aldcroft, T., Davis, M., Ginsburg, A.,
  {Price-Whelan}, A.~M., Kerzendorf, W.~E., Conley, A., Crighton, N., Barbary,
  K., Muna, D., Ferguson, H., Grollier, F., Parikh, M.~M., Nair, P.~H., Unther,
  H.~M., Deil, C., Woillez, J., Conseil, S., Kramer, R., Turner, J. E.~H.,
  Singer, L., Fox, R., Weaver, B.~A., Zabalza, V., Edwards, Z.~I.,
  Azalee~Bostroem, K., Burke, D.~J., Casey, A.~R., Crawford, S.~M., Dencheva,
  N., Ely, J., Jenness, T., Labrie, K., Lim, P.~L., Pierfederici, F., Pontzen,
  A., Ptak, A., Refsdal, B., Servillat, M., and Streicher, O.
\newblock Astropy: {{A}} community {{Python}} package for astronomy.
\newblock \emph{Astronomy and Astrophysics}, 558:\penalty0 A33, October 2013.
\newblock \doi{10.1051/0004-6361/201322068}.
\newblock URL \url{http://dx.doi.org/10.1051/0004-6361/201322068}.

\bibitem[{Astropy Collaboration} et~al.(2018){Astropy Collaboration},
  {Price-Whelan}, Sip{\H o}cz, G{\"u}nther, Lim, Crawford, Conseil, Shupe,
  Craig, Dencheva, Ginsburg, VanderPlas, Bradley, {P{\'e}rez-Su{\'a}rez}, {de
  Val-Borro}, Aldcroft, Cruz, Robitaille, Tollerud, Ardelean, Babej, Bach,
  Bachetti, Bakanov, Bamford, Barentsen, Barmby, Baumbach, Berry, Biscani,
  Boquien, Bostroem, Bouma, Brammer, Bray, Breytenbach, Buddelmeijer, Burke,
  Calderone, Cano~Rodr{\'\i}guez, Cara, Cardoso, Cheedella, Copin, Corrales,
  Crichton, D'Avella, Deil, Depagne, Dietrich, Donath, Droettboom, Earl, Erben,
  Fabbro, Ferreira, Finethy, Fox, Garrison, Gibbons, Goldstein, Gommers, Greco,
  Greenfield, Groener, Grollier, Hagen, Hirst, Homeier, Horton, Hosseinzadeh,
  Hu, Hunkeler, Ivezi{\'c}, Jain, Jenness, Kanarek, Kendrew, Kern, Kerzendorf,
  Khvalko, King, Kirkby, Kulkarni, Kumar, Lee, Lenz, Littlefair, Ma, Macleod,
  Mastropietro, McCully, Montagnac, Morris, Mueller, Mumford, Muna, Murphy,
  Nelson, Nguyen, Ninan, N{\"o}the, Ogaz, Oh, Parejko, Parley, Pascual, Patil,
  Patil, Plunkett, Prochaska, Rastogi, Reddy~Janga, Sabater, Sakurikar,
  Seifert, Sherbert, {Sherwood-Taylor}, Shih, Sick, Silbiger, Singanamalla,
  Singer, Sladen, Sooley, Sornarajah, Streicher, Teuben, Thomas, Tremblay,
  Turner, Terr{\'o}n, {van Kerkwijk}, {de la Vega}, Watkins, Weaver, Whitmore,
  Woillez, Zabalza, and {Astropy Contributors}]{2018AJ....156..123A}
{Astropy Collaboration}, {Price-Whelan}, A.~M., Sip{\H o}cz, B.~M.,
  G{\"u}nther, H.~M., Lim, P.~L., Crawford, S.~M., Conseil, S., Shupe, D.~L.,
  Craig, M.~W., Dencheva, N., Ginsburg, A., VanderPlas, J.~T., Bradley, L.~D.,
  {P{\'e}rez-Su{\'a}rez}, D., {de Val-Borro}, M., Aldcroft, T.~L., Cruz, K.~L.,
  Robitaille, T.~P., Tollerud, E.~J., Ardelean, C., Babej, T., Bach, Y.~P.,
  Bachetti, M., Bakanov, A.~V., Bamford, S.~P., Barentsen, G., Barmby, P.,
  Baumbach, A., Berry, K.~L., Biscani, F., Boquien, M., Bostroem, K.~A., Bouma,
  L.~G., Brammer, G.~B., Bray, E.~M., Breytenbach, H., Buddelmeijer, H., Burke,
  D.~J., Calderone, G., Cano~Rodr{\'\i}guez, J.~L., Cara, M., Cardoso, J.
  V.~M., Cheedella, S., Copin, Y., Corrales, L., Crichton, D., D'Avella, D.,
  Deil, C., Depagne, {\'E}., Dietrich, J.~P., Donath, A., Droettboom, M., Earl,
  N., Erben, T., Fabbro, S., Ferreira, L.~A., Finethy, T., Fox, R.~T.,
  Garrison, L.~H., Gibbons, S. L.~J., Goldstein, D.~A., Gommers, R., Greco,
  J.~P., Greenfield, P., Groener, A.~M., Grollier, F., Hagen, A., Hirst, P.,
  Homeier, D., Horton, A.~J., Hosseinzadeh, G., Hu, L., Hunkeler, J.~S.,
  Ivezi{\'c}, {\v Z}., Jain, A., Jenness, T., Kanarek, G., Kendrew, S., Kern,
  N.~S., Kerzendorf, W.~E., Khvalko, A., King, J., Kirkby, D., Kulkarni, A.~M.,
  Kumar, A., Lee, A., Lenz, D., Littlefair, S.~P., Ma, Z., Macleod, D.~M.,
  Mastropietro, M., McCully, C., Montagnac, S., Morris, B.~M., Mueller, M.,
  Mumford, S.~J., Muna, D., Murphy, N.~A., Nelson, S., Nguyen, G.~H., Ninan,
  J.~P., N{\"o}the, M., Ogaz, S., Oh, S., Parejko, J.~K., Parley, N., Pascual,
  S., Patil, R., Patil, A.~A., Plunkett, A.~L., Prochaska, J.~X., Rastogi, T.,
  Reddy~Janga, V., Sabater, J., Sakurikar, P., Seifert, M., Sherbert, L.~E.,
  {Sherwood-Taylor}, H., Shih, A.~Y., Sick, J., Silbiger, M.~T., Singanamalla,
  S., Singer, L.~P., Sladen, P.~H., Sooley, K.~A., Sornarajah, S., Streicher,
  O., Teuben, P., Thomas, S.~W., Tremblay, G.~R., Turner, J. E.~H., Terr{\'o}n,
  V., {van Kerkwijk}, M.~H., {de la Vega}, A., Watkins, L.~L., Weaver, B.~A.,
  Whitmore, J.~B., Woillez, J., Zabalza, V., and {Astropy Contributors}.
\newblock The {{Astropy Project}}: {{Building}} an {{Open-science Project}} and
  {{Status}} of the v2.0 {{Core Package}}.
\newblock \emph{The Astronomical Journal}, 156:\penalty0 123, September 2018.
\newblock ISSN 0004-6256.
\newblock \doi{10.3847/1538-3881/aabc4f}.
\newblock URL \url{https://ui.adsabs.harvard.edu/abs/2018AJ....156..123A}.

\bibitem[Blundell et~al.(2015)Blundell, Cornebise, Kavukcuoglu, and
  Wierstra]{blundell2015weight}
Blundell, C., Cornebise, J., Kavukcuoglu, K., and Wierstra, D.
\newblock Weight uncertainty in neural network.
\newblock In \emph{International Conference on Machine Learning}, pp.\
  1613--1622. PMLR, 2015.

\bibitem[Cranmer et~al.(2020)Cranmer, Brehmer, and Louppe]{cranmer_sbi2020}
Cranmer, K., Brehmer, J., and Louppe, G.
\newblock The frontier of simulation-based inference.
\newblock \emph{Proceedings of the National Academy of Sciences}, 117\penalty0
  (48):\penalty0 30055--30062, 2020.
\newblock \doi{10.1073/pnas.1912789117}.

\bibitem[Gal \& Ghahramani(2016)Gal and Ghahramani]{gal2016dropout}
Gal, Y. and Ghahramani, Z.
\newblock Dropout as a bayesian approximation: Representing model uncertainty
  in deep learning.
\newblock In \emph{international conference on machine learning}, pp.\
  1050--1059. PMLR, 2016.

\bibitem[Gneiting \& Raftery(2007)Gneiting and Raftery]{gneiting2007strictly}
Gneiting, T. and Raftery, A.~E.
\newblock Strictly proper scoring rules, prediction, and estimation.
\newblock \emph{Journal of the American statistical Association}, 102\penalty0
  (477):\penalty0 359--378, 2007.

\bibitem[Goodfellow et~al.(2015)Goodfellow, Shlens, and
  Szegedy]{Goodfellow_2015}
Goodfellow, I.~J., Shlens, J., and Szegedy, C.
\newblock Explaining and harnessing adversarial examples.
\newblock \emph{ICLR}, 2015.

\bibitem[Guo et~al.(2017)Guo, Pleiss, Sun, and Weinberger]{guo2017calibration}
Guo, C., Pleiss, G., Sun, Y., and Weinberger, K.~Q.
\newblock On calibration of modern neural networks.
\newblock In \emph{International Conference on Machine Learning}, pp.\
  1321--1330. PMLR, 2017.

\bibitem[Harris et~al.(2020)Harris, Millman, {van der Walt}, Gommers, Virtanen,
  Cournapeau, Wieser, Taylor, Berg, Smith, Kern, Picus, Hoyer, {van Kerkwijk},
  Brett, Haldane, {del R{\'\i}o}, Wiebe, Peterson, {G{\'e}rard-Marchant},
  Sheppard, Reddy, Weckesser, Abbasi, Gohlke, and
  Oliphant]{2020Natur.585..357H}
Harris, C.~R., Millman, K.~J., {van der Walt}, S.~J., Gommers, R., Virtanen,
  P., Cournapeau, D., Wieser, E., Taylor, J., Berg, S., Smith, N.~J., Kern, R.,
  Picus, M., Hoyer, S., {van Kerkwijk}, M.~H., Brett, M., Haldane, A., {del
  R{\'\i}o}, J.~F., Wiebe, M., Peterson, P., {G{\'e}rard-Marchant}, P.,
  Sheppard, K., Reddy, T., Weckesser, W., Abbasi, H., Gohlke, C., and Oliphant,
  T.~E.
\newblock Array programming with {{NumPy}}.
\newblock \emph{Nature}, 585:\penalty0 357--362, September 2020.
\newblock ISSN 0028-0836.
\newblock \doi{10.1038/s41586-020-2649-2}.
\newblock URL \url{https://ui.adsabs.harvard.edu/abs/2020Natur.585..357H}.

\bibitem[He et~al.(2016)He, Zhang, Ren, and Sun]{he2016deep}
He, K., Zhang, X., Ren, S., and Sun, J.
\newblock Deep residual learning for image recognition.
\newblock In \emph{Proceedings of the IEEE conference on computer vision and
  pattern recognition}, pp.\  770--778, 2016.

\bibitem[Kerzendorf et~al.(2022{\natexlab{a}})Kerzendorf, Sim, Vogl,
  Williamson, P{\'a}ssaro, Fl{\"o}rs, Camacho, Jan{\v c}auskas, Harpole,
  N{\"o}bauer, Lietzau, Mishin, Tsamis, Boyle, Shingles, Gupta, Desai, Klauser,
  Beaujean, Suban-Loewen, Heringer, Barna, Gautam, Fullard, Cawley, Singhal,
  Smith, Barbosa, Sondhi, Patel, Varanasi, Gillanders, Arya, O'Brien, Eweis,
  Reinecke, Bylund, Bentil, Savel, Yu, Eguren, Alam, Magee, Livneh,
  Rajagopalan, Mishra, Reichenbach, Jain, Floers, Brar, Singh, Talegaonkar,
  Kowalski, Selsing, Sofiatti, Aggarwal, Sarafina, Patra, Singh~Rathore, Patel,
  Sharma, Gupta, Wahi, Sandler, Volodin, Dasgupta, Yap, Kharkar, Nayak~U,
  Kolliboyina, and Kumar]{kerzendorf_wolfgang_2022_6430631}
Kerzendorf, W., Sim, S., Vogl, C., Williamson, M., P{\'a}ssaro, E., Fl{\"o}rs,
  A., Camacho, Y., Jan{\v c}auskas, V., Harpole, A., N{\"o}bauer, U., Lietzau,
  S., Mishin, M., Tsamis, F., Boyle, A., Shingles, L., Gupta, V., Desai, K.,
  Klauser, M., Beaujean, F., Suban-Loewen, A., Heringer, E., Barna, B., Gautam,
  G., Fullard, A., Cawley, K., Singhal, J., Smith, I., Barbosa, T., Sondhi, D.,
  Patel, M., Varanasi, K., Gillanders, J., Arya, A., O'Brien, J., Eweis, Y.,
  Reinecke, M., Bylund, T., Bentil, L., Savel, A., Yu, J., Eguren, J., Alam,
  A., Magee, M., Livneh, R., Rajagopalan, S., Mishra, S., Reichenbach, J.,
  Jain, R., Floers, A., Brar, A., Singh, S., Talegaonkar, C., Kowalski, N.,
  Selsing, J., Sofiatti, C., Aggarwal, Y., Sarafina, N., Patra, N.,
  Singh~Rathore, P., Patel, P., Sharma, S., Gupta, S., Wahi, U., Sandler, M.,
  Volodin, D., Dasgupta, D., Yap, K., Kharkar, A., Nayak~U, A., Kolliboyina,
  C., and Kumar, A.
\newblock tardis-sn/tardis: Tardis v2022.4.1, April 2022{\natexlab{a}}.
\newblock URL \url{https://doi.org/10.5281/zenodo.6430631}.

\bibitem[Kerzendorf et~al.(2022{\natexlab{b}})Kerzendorf, Sim, Vogl,
  Williamson, P{\'a}ssaro, Fl{\"o}rs, Camacho, Jan{\v c}auskas, Harpole,
  N{\"o}bauer, Lietzau, Mishin, Tsamis, Boyle, Shingles, Gupta, Desai, Klauser,
  Beaujean, Suban-Loewen, Heringer, Barna, Gautam, Fullard, Cawley, Smith,
  Singhal, Arya, Sondhi, Barbosa, Yu, Patel, O'Brien, Varanasi, Gillanders,
  Savel, Reinecke, Eweis, Bylund, Bentil, Eguren, Alam, Bartnik, Magee,
  Shields, Livneh, Rajagopalan, Chitchyan, Mishra, Reichenbach, Jain, Floers,
  Brar, Singh, Selsing, Sofiatti, Talegaonkar, Bot, Kowalski, Yap, Patel,
  Sharma, Prasad, Venkat, Dasgupta, Zaheer, Gupta, Volodin, Patra,
  Singh~Rathore, Lemoine, Sarafina, Kolliboyina, Sandler, Nayak~U, Aggarwal,
  Kumar, Holas, Kharkar, kumar, and Wahi]{kerzendorf_wolfgang_2022_6527890}
Kerzendorf, W., Sim, S., Vogl, C., Williamson, M., P{\'a}ssaro, E., Fl{\"o}rs,
  A., Camacho, Y., Jan{\v c}auskas, V., Harpole, A., N{\"o}bauer, U., Lietzau,
  S., Mishin, M., Tsamis, F., Boyle, A., Shingles, L., Gupta, V., Desai, K.,
  Klauser, M., Beaujean, F., Suban-Loewen, A., Heringer, E., Barna, B., Gautam,
  G., Fullard, A., Cawley, K., Smith, I., Singhal, J., Arya, A., Sondhi, D.,
  Barbosa, T., Yu, J., Patel, M., O'Brien, J., Varanasi, K., Gillanders, J.,
  Savel, A., Reinecke, M., Eweis, Y., Bylund, T., Bentil, L., Eguren, J., Alam,
  A., Bartnik, M., Magee, M., Shields, J., Livneh, R., Rajagopalan, S.,
  Chitchyan, S., Mishra, S., Reichenbach, J., Jain, R., Floers, A., Brar, A.,
  Singh, S., Selsing, J., Sofiatti, C., Talegaonkar, C., Bot, T., Kowalski, N.,
  Yap, K., Patel, P., Sharma, S., Prasad, S., Venkat, S., Dasgupta, D., Zaheer,
  M., Gupta, S., Volodin, D., Patra, N., Singh~Rathore, P., Lemoine, T.,
  Sarafina, N., Kolliboyina, C., Sandler, M., Nayak~U, A., Aggarwal, Y., Kumar,
  A., Holas, A., Kharkar, A., kumar, a., and Wahi, U.
\newblock tardis-sn/tardis: Tardis v2022.05.08, May 2022{\natexlab{b}}.
\newblock URL \url{https://doi.org/10.5281/zenodo.6527890}.

\bibitem[{Kerzendorf} \& {Sim}(2014){Kerzendorf} and
  {Sim}]{2014MNRAS.440..387K}
{Kerzendorf}, W.~E. and {Sim}, S.~A.
\newblock {A spectral synthesis code for rapid modelling of supernovae}.
\newblock \emph{\mnras}, 440:\penalty0 387--404, May 2014.
\newblock \doi{10.1093/mnras/stu055}.

\bibitem[Kerzendorf et~al.(2021)Kerzendorf, Vogl, Buchner, Contardo,
  Williamson, and van~der Smagt]{kerzendorf2021dalek}
Kerzendorf, W.~E., Vogl, C., Buchner, J., Contardo, G., Williamson, M., and
  van~der Smagt, P.
\newblock Dalek: A deep learning emulator for tardis.
\newblock \emph{The Astrophysical Journal Letters}, 910\penalty0 (2):\penalty0
  L23, 2021.

\bibitem[Lakshminarayanan et~al.(2017)Lakshminarayanan, Pritzel, and
  Blundell]{NIPS2017_9ef2ed4b}
Lakshminarayanan, B., Pritzel, A., and Blundell, C.
\newblock Simple and scalable predictive uncertainty estimation using deep
  ensembles.
\newblock In \emph{Advances in Neural Information Processing Systems},
  volume~30. Curran Associates, Inc., 2017.

\bibitem[Liu et~al.(2020)Liu, Lin, Padhy, Tran, Bedrax~Weiss, and
  Lakshminarayanan]{NEURIPS2020_543e8374}
Liu, J., Lin, Z., Padhy, S., Tran, D., Bedrax~Weiss, T., and Lakshminarayanan,
  B.
\newblock Simple and principled uncertainty estimation with deterministic deep
  learning via distance awareness.
\newblock In Larochelle, H., Ranzato, M., Hadsell, R., Balcan, M.~F., and Lin,
  H. (eds.), \emph{Advances in Neural Information Processing Systems},
  volume~33, pp.\  7498--7512. Curran Associates, Inc., 2020.

\bibitem[{O'Brien} et~al.(2021){O'Brien}, {Kerzendorf}, {Fullard},
  {Williamson}, {Pakmor}, {Buchner}, {Hachinger}, {Vogl}, {Gillanders},
  {Fl{\"o}rs}, and {van der Smagt}]{obrien21}
{O'Brien}, J.~T., {Kerzendorf}, W.~E., {Fullard}, A., {Williamson}, M.,
  {Pakmor}, R., {Buchner}, J., {Hachinger}, S., {Vogl}, C., {Gillanders},
  J.~H., {Fl{\"o}rs}, A., and {van der Smagt}, P.
\newblock {Probabilistic Reconstruction of Type Ia Supernova SN 2002bo}.
\newblock \emph{ApJ Letter}, 916\penalty0 (2):\penalty0 L14, August 2021.
\newblock \doi{10.3847/2041-8213/ac1173}.

\bibitem[Ovadia et~al.(2019)Ovadia, Fertig, Ren, Nado, Sculley, Nowozin,
  Dillon, Lakshminarayanan, and Snoek]{NEURIPS2019_8558cb40}
Ovadia, Y., Fertig, E., Ren, J., Nado, Z., Sculley, D., Nowozin, S., Dillon,
  J., Lakshminarayanan, B., and Snoek, J.
\newblock Can you trust your model\textquotesingle s uncertainty? evaluating
  predictive uncertainty under dataset shift.
\newblock In \emph{Advances in Neural Information Processing Systems},
  volume~32. Curran Associates, Inc., 2019.

\bibitem[Paszke et~al.(2019)Paszke, Gross, Massa, Lerer, Bradbury, Chanan,
  Killeen, Lin, Gimelshein, Antiga, Desmaison, Kopf, Yang, DeVito, Raison,
  Tejani, Chilamkurthy, Steiner, Fang, Bai, and Chintala]{NEURIPS2019_9015}
Paszke, A., Gross, S., Massa, F., Lerer, A., Bradbury, J., Chanan, G., Killeen,
  T., Lin, Z., Gimelshein, N., Antiga, L., Desmaison, A., Kopf, A., Yang, E.,
  DeVito, Z., Raison, M., Tejani, A., Chilamkurthy, S., Steiner, B., Fang, L.,
  Bai, J., and Chintala, S.
\newblock Pytorch: An imperative style, high-performance deep learning library.
\newblock In \emph{Advances in Neural Information Processing Systems 32}, pp.\
  8024--8035. Curran Associates, Inc., 2019.

\bibitem[Pedregosa et~al.(2011)Pedregosa, Varoquaux, Gramfort, Michel, Thirion,
  Grisel, Blondel, Prettenhofer, Weiss, Dubourg, et~al.]{pedregosa2011scikit}
Pedregosa, F., Varoquaux, G., Gramfort, A., Michel, V., Thirion, B., Grisel,
  O., Blondel, M., Prettenhofer, P., Weiss, R., Dubourg, V., et~al.
\newblock Scikit-learn: Machine learning in python.
\newblock \emph{the Journal of machine Learning research}, 12:\penalty0
  2825--2830, 2011.

\bibitem[Reback et~al.(2022)Reback, jbrockmendel, McKinney, den Bossche,
  Roeschke, Augspurger, Hawkins, Cloud, gfyoung, Sinhrks, Hoefler, Klein,
  Petersen, Tratner, She, Ayd, Naveh, Darbyshire, Shadrach, Garcia, Schendel,
  Hayden, Saxton, Gorelli, Li, W{\"o}rtwein, Zeitlin, Jancauskas, McMaster, and
  Li]{jeff_reback_2022_6702671}
Reback, J., jbrockmendel, McKinney, W., den Bossche, J.~V., Roeschke, M.,
  Augspurger, T., Hawkins, S., Cloud, P., gfyoung, Sinhrks, Hoefler, P., Klein,
  A., Petersen, T., Tratner, J., She, C., Ayd, W., Naveh, S., Darbyshire, J.,
  Shadrach, R., Garcia, M., Schendel, J., Hayden, A., Saxton, D., Gorelli,
  M.~E., Li, F., W{\"o}rtwein, T., Zeitlin, M., Jancauskas, V., McMaster, A.,
  and Li, T.
\newblock pandas-dev/pandas: Pandas 1.4.3, June 2022.
\newblock URL \url{https://doi.org/10.5281/zenodo.6702671}.

\bibitem[Szegedy et~al.(2014)Szegedy, Zaremba, Sutskever, Bruna, Erhan,
  Goodfellow, and Fergus]{Szegedy}
Szegedy, C., Zaremba, W., Sutskever, I., Bruna, J., Erhan, D., Goodfellow, I.,
  and Fergus, R.
\newblock Intriguing properties of neural networks.
\newblock In \emph{International Conference on Learning Representations}, 2014.

\bibitem[{The Astropy Collaboration} et~al.(2022){The Astropy Collaboration},
  {Price-Whelan}, Lian~Lim, Earl, Starkman, Bradley, Shupe, Patil, Corrales,
  Brasseur, N{\"o}the, Donath, Tollerud, Morris, Ginsburg, Vaher, Weaver,
  Tocknell, Jamieson, {van Kerkwijk}, Robitaille, Merry, Bachetti, G{\"u}nther,
  Aldcroft, {Alvarado-Montes}, Archibald, B{\'o}di, Bapat, Barentsen,
  Baz{\'a}n, Biswas, Boquien, Burke, Cara, Cara, E~Conroy, Conseil, Craig,
  Cross, Cruz, D'Eugenio, Dencheva, Devillepoix, Dietrich, Davis~Eigenbrot,
  Erben, Ferreira, {Foreman-Mackey}, Fox, Freij, Garg, Geda, Glattly,
  Gondhalekar, Gordon, Grant, Greenfield, Groener, Guest, Gurovich, Handberg,
  Hart, {Hatfield-Dodds}, Homeier, Hosseinzadeh, Jenness, Jones, Joseph,
  Bryce~Kalmbach, Karamehmetoglu, Ka{\l}uszy{\'n}ski, Kelley, Kern, Kerzendorf,
  Koch, Kulumani, Lee, Ly, Ma, MacBride, Maljaars, Muna, Murphy, Norman,
  O'Steen, Oman, Pacifici, Pascual, {Pascual-Granado}, Patil, Perren,
  Pickering, Rastogi, Roulston, Ryan, Rykoff, Sabater, Sakurikar, Salgado,
  Sanghi, Saunders, Savchenko, Schwardt, {Seifert-Eckert}, Shih, Shrey~Jain,
  Shukla, Sick, Simpson, Singanamalla, Singer, Singhal, Sinha, Sip{\H o}cz,
  Spitler, Stansby, Streicher, {\v S}umak, Swinbank, Taranu, Tewary, Tremblay,
  {de Val-Borro}, Van~Kooten, Vasovi{\'c}, Verma, Cardoso, Williams, Wilson,
  Winkel, {Wood-Vasey}, Xue, Yoachim, ZHANG, and Zonca]{2022arXiv220614220T}
{The Astropy Collaboration}, {Price-Whelan}, A.~M., Lian~Lim, P., Earl, N.,
  Starkman, N., Bradley, L., Shupe, D.~L., Patil, A.~A., Corrales, L.,
  Brasseur, C.~E., N{\"o}the, M., Donath, A., Tollerud, E., Morris, B.~M.,
  Ginsburg, A., Vaher, E., Weaver, B.~A., Tocknell, J., Jamieson, W., {van
  Kerkwijk}, M.~H., Robitaille, T.~P., Merry, B., Bachetti, M., G{\"u}nther,
  H.~M., Aldcroft, T.~L., {Alvarado-Montes}, J.~A., Archibald, A.~M., B{\'o}di,
  A., Bapat, S., Barentsen, G., Baz{\'a}n, J., Biswas, M., Boquien, M., Burke,
  D.~J., Cara, D., Cara, M., E~Conroy, K., Conseil, S., Craig, M.~W., Cross,
  R.~M., Cruz, K.~L., D'Eugenio, F., Dencheva, N., Devillepoix, H. A.~R.,
  Dietrich, J.~P., Davis~Eigenbrot, A., Erben, T., Ferreira, L.,
  {Foreman-Mackey}, D., Fox, R., Freij, N., Garg, S., Geda, R., Glattly, L.,
  Gondhalekar, Y., Gordon, K.~D., Grant, D., Greenfield, P., Groener, A.~M.,
  Guest, S., Gurovich, S., Handberg, R., Hart, A., {Hatfield-Dodds}, Z.,
  Homeier, D., Hosseinzadeh, G., Jenness, T., Jones, C.~K., Joseph, P.,
  Bryce~Kalmbach, J., Karamehmetoglu, E., Ka{\l}uszy{\'n}ski, M., Kelley, M.
  S.~P., Kern, N., Kerzendorf, W.~E., Koch, E.~W., Kulumani, S., Lee, A., Ly,
  C., Ma, Z., MacBride, C., Maljaars, J.~M., Muna, D., Murphy, N.~A., Norman,
  H., O'Steen, R., Oman, K.~A., Pacifici, C., Pascual, S., {Pascual-Granado},
  J., Patil, R.~R., Perren, G.~I., Pickering, T.~E., Rastogi, T., Roulston,
  B.~R., Ryan, D.~F., Rykoff, E.~S., Sabater, J., Sakurikar, P., Salgado, J.,
  Sanghi, A., Saunders, N., Savchenko, V., Schwardt, L., {Seifert-Eckert}, M.,
  Shih, A.~Y., Shrey~Jain, A., Shukla, G., Sick, J., Simpson, C., Singanamalla,
  S., Singer, L.~P., Singhal, J., Sinha, M., Sip{\H o}cz, B.~M., Spitler,
  L.~R., Stansby, D., Streicher, O., {\v S}umak, J., Swinbank, J.~D., Taranu,
  D.~S., Tewary, N., Tremblay, G.~R., {de Val-Borro}, M., Van~Kooten, S.~J.,
  Vasovi{\'c}, Z., Verma, S., Cardoso, J. V. d.~M., Williams, P. K.~G., Wilson,
  T.~J., Winkel, B., {Wood-Vasey}, W.~M., Xue, R., Yoachim, P., ZHANG, C., and
  Zonca, A.
\newblock The {{Astropy Project}}: {{Sustaining}} and {{Growing}} a
  {{Community-oriented Open-source Project}} and the {{Latest Major Release}}
  (v5.0) of the {{Core Package}}, June 2022.
\newblock URL \url{https://ui.adsabs.harvard.edu/abs/2022arXiv220614220T}.

\bibitem[Van~Amersfoort et~al.(2020)Van~Amersfoort, Smith, Teh, and
  Gal]{van2020uncertainty}
Van~Amersfoort, J., Smith, L., Teh, Y.~W., and Gal, Y.
\newblock Uncertainty estimation using a single deep deterministic neural
  network.
\newblock In \emph{International Conference on Machine Learning}, pp.\
  9690--9700. PMLR, 2020.

\bibitem[van~der Smagt \& Hirzinger(1998)van~der Smagt and
  Hirzinger]{van1998solving}
van~der Smagt, P. and Hirzinger, G.
\newblock Solving the ill-conditioning in neural network learning.
\newblock \emph{Lecture notes in computer science}, pp.\  193--206, 1998.

\bibitem[{Vogl} et~al.(2019){Vogl}, {Sim}, {Noebauer}, {Kerzendorf}, and
  {Hillebrandt}]{2019A&A...621A..29V}
{Vogl}, C., {Sim}, S.~A., {Noebauer}, U.~M., {Kerzendorf}, W.~E., and
  {Hillebrandt}, W.
\newblock {Spectral modeling of type II supernovae. I. Dilution factors}.
\newblock \emph{\aap}, 621:\penalty0 A29, Jan 2019.
\newblock \doi{10.1051/0004-6361/201833701}.

\bibitem[{Vogl, C.} et~al.(2020){Vogl, C.}, {Kerzendorf, W. E.}, {Sim, S. A.},
  {Noebauer, U. M.}, {Lietzau, S.}, and {Hillebrandt, W.}]{refId0}
{Vogl, C.}, {Kerzendorf, W. E.}, {Sim, S. A.}, {Noebauer, U. M.}, {Lietzau,
  S.}, and {Hillebrandt, W.}
\newblock Spectral modeling of type ii supernovae - ii. a machine-learning
  approach to quantitative spectroscopic analysis.
\newblock \emph{A\&A}, 633:\penalty0 A88, 2020.
\newblock \doi{10.1051/0004-6361/201936137}.

\end{thebibliography}
\bibliographystyle{icml2022.bst}
\end{document}